\documentclass[letterpaper, 10 pt, conference]{ieeeconf}  

\IEEEoverridecommandlockouts                              

\usepackage{times} 
\usepackage[sorting=none, maxcitenames=1, maxbibnames=99, style=numeric-comp]{biblatex}
\AtBeginBibliography{\small}
\usepackage{graphicx}
\usepackage[font=small]{caption}
\usepackage{amsmath}
\usepackage{amssymb}
\usepackage{hyperref}
\hypersetup{colorlinks,linkcolor={black},citecolor={magenta},urlcolor={magenta}}

\usepackage{algorithm} 
\usepackage{algpseudocode} 

\title{\LARGE \bf
On the Feasibility of Learning Finger-gaiting In-hand Manipulation\\with Intrinsic Sensing
}

\author{Gagan Khandate, Maximilian Haas-Heger, Matei Ciocarlie \\ 
\thanks{When this work was performed, all authors were with Columbia University, New York, NY 10027, USA. Gagan Khandate: Department of Computer Science; Maxmillian Haas-Heger and Matei Ciocarlie: Department of Mechanical Engineering. Emails: {\tt\small gagank@cs.columbia.edu}, {\tt\small m.haas@columbia.edu}, {\tt\small matei.ciocarlie@columbia.edu}}
\thanks{This work was supported in part by NSF grant CMMI-1734557, ONR grant N00014-19-1-2062, and by an Amazon Research Award.}
}
\begin{document}


\maketitle
\thispagestyle{empty}
\pagestyle{empty}

\begin{abstract}
Finger-gaiting manipulation is an important skill to achieve large-angle in-hand re-orientation of objects. However, achieving these gaits with arbitrary orientations of the hand is challenging due to the unstable nature of the task. In this work, we use model-free reinforcement learning (RL) to learn finger-gaiting only via precision grasps and demonstrate finger-gaiting for rotation about an axis using only on-board proprioceptive and tactile feedback. To tackle the inherent instability of precision grasping, we propose the use of initial state distributions that enable effective exploration of the state space. Our method can learn finger gaiting with better sample complexity than the state-of-the-art. The policies we obtain are robust to noise and perturbations, and transfer to novel objects. Videos can be found at \href{https://roamlab.github.io/learnfg/}{https://roamlab.github.io/learnfg/}
\end{abstract}

\section{Introduction}
Dexterous in-hand manipulation \cite{Okamura2000-in} is the ability to move a grasped object into a desired pose within the hand. Humans routinely use in-hand manipulation to perform tasks such as re-orienting a tool from the initial grasp into a useful pose, securing a better grasp on the object, exploring the shape of an unknown object, etc. Thus, robotic in-hand manipulation is an important step towards the general goal of manipulating objects in cluttered and unstructured environments such as in a kitchen or a warehouse. However, versatile in-hand manipulation remains a long standing challenge.

A whole spectrum of methods have been considered for in-hand manipulation; online trajectory optimization methods \cite{Tassa2012-hx} and model-free deep reinforcement learning (RL) \cite{Schulman2017-fr} stand out for highly actuated dexterous hands. Model-based online trajectory optimization methods have succeeded in generating complex behaviors for dexterous robotic manipulation, but not for finger-gaiting as these tasks fatally exacerbate their limitations: transient contacts introduce large non-linearities in the model, which also depends on hard-to-model contact properties.

While RL has been successful in demonstrating diverse in-hand manipulation skills both in simulation and on real hands \cite{OpenAI2018-bx}, the policies obtained are object-centric and require large training times. In many cases, these policies do not transfer to arbitrary orientations of the hand as they expect the palm to support the object during manipulation \textemdash a consequence of the policies being trained in a palm-up hand orientation, which simplifies training. In other cases, the policies require extensive external sensing involving multi-camera systems to track the fingers and/or the object, systems that are hard to deploy outside the lab.

Tactile feedback has the potential to enable robust and generalizable in-hand manipulation \cite{Li2020-mc} and to eliminate the need for external sensing. However, integrating tactile feedback with RL is a challenge of its own. Besides the general difficulty of simulating the transduction involved, tactile feedback is often high dimensional which can prohibitively drive up the number of training samples required. Hence, prior works using RL for in-hand manipulation either avoid using tactile feedback altogether, or consider tasks requiring fewer training samples where it is feasible to learn directly on a real hand.

\begin{figure}[t]
  \centering
  \includegraphics[clip, width=\columnwidth]{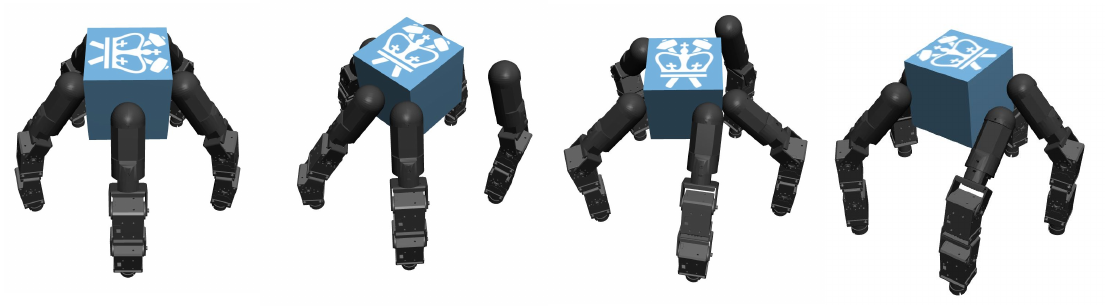}
  \caption{A learned finger-gaiting policy that can continuously re-orient the target object about the hand z-axis. The policy only uses sensing modalities intrinsic to the hand (such as touch and proprioception), and does not require explicit object pose information from external sensors.}
\end{figure}

We too use RL, but focus on learning finger-gaiting (manipulation involving finger substitution and re-grasping) and finger-pivoting (manipulation involving the object in hinge-grasp) skills. Both skills are important towards enabling large-angle in-hand object re-orientation: achieving an arbitrarily large rotation of the grasped object around a given axis, up to or even exceeding a full revolution. Such a task is generally not achievable by in-grasp manipulation (i.e. without breaking the contacts of the original grasp) and requires finger-gaiting or finger-pivoting (i.e. breaking and re-establishing contacts during manipulation); these are not restricted by the kinematic constraints of the hand and can achieve potentially limitless object re-orientation.

We are interested in achieving these skills exclusively through using fingertip grasps (i.e precision in-hand manipulation \cite{Michelman1998-ye}) without requiring the presence of the palm underneath the object, which enables the policies to be used in arbitrary orientations of the hand. However, the task of learning to manipulate only via such precision grasps is a significantly harder problem: action randomization, responsible for exploration in RL, often fails as the hand can easily drop the object.

Furthermore, we would like to circumvent the need for cumbersome external sensing by only using internal sensing in achieving these skills. The challenge here is that the absence of external sensing implies we do not have information regarding the object such as its global shape and pose. However, we posit that internal sensing by itself can provide object information sufficient towards our goal. 

We set out to determine if we can even achieve finger-gaiting and finger-pivoting skills purely through intrinsic sensing in simulation, where we evaluate both proprioceptive feedback and tactile feedback. To this end, we consider the task of continuous object re-orientation about a given axis, aiming to learn finger-gaiting and finger-pivoting without object pose information. With this approach, we hope to learn policies to rotate object about cardinal axes and combine them for arbitrary in-hand object re-orientation. To overcome challenges in exploration, we propose collecting training trajectories starting from a wide range of grasps sampled from appropriately designed initial state distributions as an alternative exploration mechanism.

We summarize the contributions of this work as follows:

\begin{enumerate}
\item We learn finger-gaiting and finger-pivoting policies that can achieve large angle in-hand re-orientation of a range of simulated objects. Our policies learn to grasp and manipulate only via precision fingertip grasps using a highly dexterous and fully actuated hand, allowing us to keep the object in a stable grasp without the need for passive support at any instance during manipulation.
\item We are the first to achieve these skills by making use of only intrinsic sensing such as proprioception and touch, while also generalizing to multiple object shapes.
\item We present an exhaustive analysis of the importance of different internal sensor feedback for learning finger-gaiting and finger-pivoting policies in a simulated environment using our approach.
\end{enumerate}
\section{Related Work}
Early model-based work on finger-gaiting \cite{Leveroni1996-iy}\cite{ Han1998-xj} \cite{Platt2004-nr} \cite{Saut2007-su} and finger-pivoting \cite{Omata1996-hp} generally make simplifying assumptions such as 2D manipulation, accurate models, and smooth object geometries which limit their versatility. More recently, \textcite{Fan2017-vz, Sundaralingam2018-zw} use model based  online optimization and demonstrate finger-gaiting in simulation. These methods either use smooth objects or require accurate kinematic models of the object, which make these methods challenging to transfer to real hands.

\textcite{OpenAI2018-bx} demonstrate finger-gaiting and finger-pivoting using RL, but as previously discussed, their policies cannot be used for arbitrary orientations of the hand. This can be achieved using only force-closed precision fingertip grasps, but learning in-hand manipulation using only these grasps is challenging with few prior work. \textcite{Li2019-kj} learn 2D re-orientation using model-based controllers to ensure grasp stability in simulation. \textcite{Veiga2020-zm} demonstrate in-hand reorientation with only fingertips but these object centric policies are limited to small re-orientations via in-grasp manipulation and still require external sensing. \textcite{Shi2020-dd} demonstrate precision finger-gaiting but only on a lightweight ball. \textcite{Morgan2021-ny} also show precision finger-gaiting but with an under-actuated hand specifically designed for this task. We consider finger-gaiting with a highly actuated hand, a harder problem due to poor sample complexity stemming from additional degrees of freedom.

Some prior work \cite{Rajeswaran2017-fw}\cite{Zhu2019-rk}\cite{Radosavovic2020-av} use human expert trajectories to improve sample complexity for dexterous manipulation.  However, these expert demonstrations are hard to obtain for precision in-hand manipulation tasks and even more so for non-anthropomorphic hands. Alternatively, model-based RL has also been considered for some in-hand manipulation tasks: \textcite{Nagabandi2020-hz} manipulate boading balls but use the palm for support; \textcite{Morgan2021-ny} learn finger-gaiting but with a task specific under-actuated hand. However, learning a reliable forward model for precision in-hand manipulation with a fully dexterous hand can be challenging. Collecting data involves random exploration, which, as we will discuss later, has difficulty exploring in this domain.

Prior work using model-free RL for manipulation rarely use tactile feedback as tactile sensing available on real hand is often high dimensional and hard to simulate \cite{OpenAI2018-bx}.  Hence, van \textcite{Van_Hoof2015-wm} propose learning directly on a real hand, but this limits them to tasks learnable on real hands. \textcite{Veiga2020-zm} learn a higher level policy through RL, while having the low level controllers exclusively deal with tactile feedback. However, this method deprives the policy from leveraging rich tactile feedback beneficial in other challenging tasks. While \textcite{Melnik2021openAItactile} show that tactile feedback improves sample complexity in such tasks, they use high-dimensional tactile feedback with full coverage that is hard to obtain on a real hand. We consider low-dimensional tactile feedback covering only the fingertips.

Contemporary to our work, \textcite{chen2021a}  show in-hand re-orientation without support surfaces that generalizes to novel objects. The policies exhibit complex dynamic behaviors including occasionally throwing the object and re-grasping it in the desired orientation. We differ from this work as our policies only use sensing that is internal to the hand, and always keep the object in a stable grasp so as to be robust to perturbation forces at all times. Furthermore, our policies require a number of training samples that is smaller by multiple orders of magnitude, a feature that we attribute to efficient exploration via appropriate initial state distributions.
\vspace{-5pt}
\section{Learning precision in-hand re-orientation}
In this work, we address two important challenges for precision in-hand re-orientation using reinforcement learning. First, we propose a hand-centric decomposition method for achieving arbitrary in-hand re-orientation in an object-agnostic fashion. Next, we identify that a key challenge of exploration for learning precision in-hand manipulation skills can be alleviated by collecting training trajectories starting at varied stable grasps. We use these grasps to design appropriate initial state distributions for training. Our approach assumes a fully-actuated and position-controlled (torque-limited) hand.

\subsection{Hand-centric decomposition}
\label{sec:decompostion}
Our aim is to push the limits on manipulation with only intrinsic sensing, and do this in a general fashion without assuming object knowledge. Thus, we do so in a hand-centric way: we learn to rotate around axes grounded in the hand frame. This means we do not need external tracking (which presumably needs to be trained for each individual object) to provide object-pose.\footnote{We note that there exist applications where specific object poses are needed at a task level, and for such cases we envision future work where a high-level object-specific tracker makes use of our hand-centric object-agnostic policies to achieve it.} We also find that rewarding angular velocity about desired axis of rotation is conducive to learning finger-gaiting and finger-pivoting policies. However, learning a single policy for any arbitrary axis is challenging as it involves learning goal-conditioned policies, which is difficult for model-free RL.


\begin{figure}[t]
    \centering
    \includegraphics[width=\columnwidth]{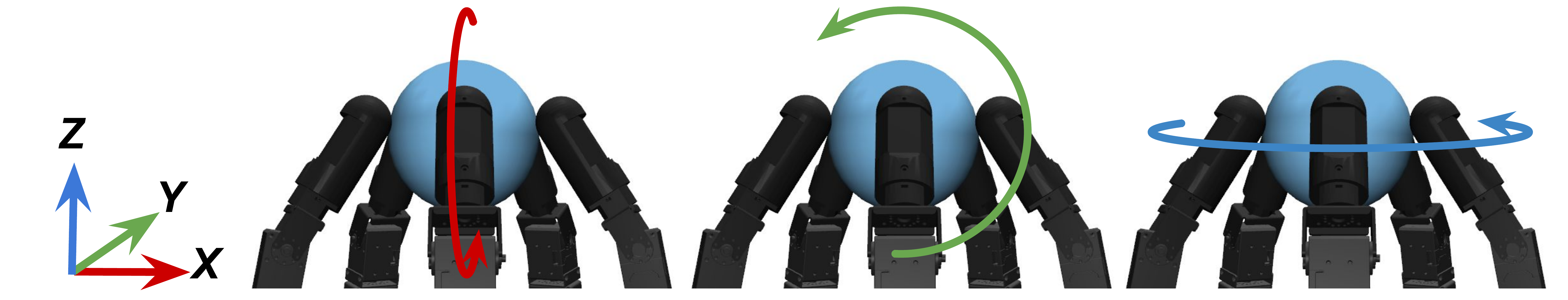}
    \caption{Hand-centric decomposition of in-hand re-orientation into re-orientation about cardinal axes.}
    \label{fig:xyz_policies}
    \vspace{-5pt}
\end{figure}

Our proposed method for large-angle arbitrary in-hand re-orientation is thus to decompose the problem of achieving arbitrary angular velocity of the object into learning separate policies about the cardinal axes as shown in Fig.~ \ref{fig:xyz_policies}. The finger-gaiting policies obtained for each axis  can then be combined in the appropriate sequence to achieve the desired change in object orientation, while side-stepping the difficulty of learning a goal-conditioned policy. 

We assume that proprioceptive sensing can provide current positions $\boldsymbol{q}$ and controller set-point positions $\boldsymbol{q}_d$. We note that the combination of desired positions and current positions can be considered as a proxy for motor forces, if the characteristics of the underlying controller are fixed. More importantly, we assume tactile sensing to provide absolute contact positions $\boldsymbol{c}^i \in \mathbb{R}^3$ and normal forces $t_{n}^i \in \mathbb{R}$ on each fingertip $i$. With known fingertip geometry, the contact normals $\hat{\mathbf{t}}_n^i \in \mathbb{R}^3$  can be derived from contact positions $\boldsymbol{c}^i$. 

Our axis-specific re-orientation policies are conditioned only on proprioceptive and tactile feedback as given by the observation vector $\mathbf{o}$:
\begin{equation}
    \mathbf{o} = [\boldsymbol{q}, \boldsymbol{q}_d, \boldsymbol{c}^1 \ldots \boldsymbol{c}^m, t_{n}^{1} \ldots t_{n}^{m}, \hat{\mathbf{t}}_{n}^{1} \ldots \hat{\mathbf{t}}_{n}^{m}]
    \label{eq:observation}
\end{equation}
where $m$ is the number of fingers. Our policies command set-point changes $\Delta \mathbf{q}_d$. 

\subsection{Learning axis-specific re-orientation}

\begin{figure}[t]
    \centering
    \includegraphics[width=0.44\columnwidth]{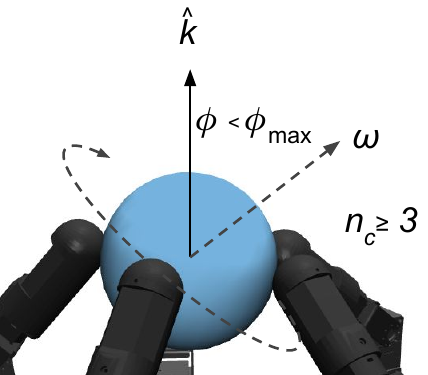}
    \caption{Learning axis conditional continuous re-orientation $\hat{\boldsymbol{k}}$. We use the component of angular velocity $\boldsymbol{\omega}$ about $\hat{\boldsymbol{k}}$ as reward when the object is in a grasp with 3 or more fingertips, i.e $n_c \geq 3$.}
    \label{fig:reward}
    \vspace{-5pt}
\end{figure}

We now describe the procedure for learning in-hand re-orientation policies for an arbitrary but fixed axis. Let $\hat{\boldsymbol{k}}$ be the desired axis of rotation. 
To learn axis-specific policy $\pi^{\hat{\boldsymbol{k}}}$ that continuously re-orients the object about the desired axis $\hat{\boldsymbol{k}}$ we use the object's angular velocity $\boldsymbol{\omega}$ along $\hat{\boldsymbol{k}}$ as reward as shown in Fig \ref{fig:reward}. However, to ensure that the policy learns to only use precision fingertip grasps to re-orient the object, we provide this reward if only fingertips are in contact with the object. In addition, we require that at least 3 fingertips are in contact with the object. Also, we encourage alignment of the object's axis of rotation with the desired axis by requiring the separation to be limited to $\phi_{max}$.

\setlength{\arraycolsep}{0.0em}
\begin{eqnarray}
r&{}={}&\min(r_{max}, {\boldsymbol{\omega}}.\hat{\boldsymbol{k}}) \ \mathbf{I}[n_c \geq 3 \land \phi \leq \phi_{max} ]\nonumber\\
&&{+}\:\min(0, {\boldsymbol{\omega}}.\hat{\mathbf{k}}) \ \mathbf{I}[n_c < 3 \vee \phi> \phi_{max} ]
\label{eq:reward}
\end{eqnarray}
\setlength{\arraycolsep}{5pt}

The reward function is described in (\ref{eq:reward}), where $n_c$ is the number of fingertip contacts and $\phi$ is the separation between the desired and current axis of rotation. Symbols $\land$, $\lor$, $\mathbf{I}$ are the logical \textit{and}, the logical \textit{or}, and indicator function, respectively. Notice that we also use reward clipping to avoid local optima and idiosyncratic behaviors. In our setup, $r_{max}$ and $\phi_{max}$ are both set to 0.5. Although the reward uses the object's angular velocity, we do not need additional sensing to measure it as we only train in simulation with the intent of zero-shot transfer to hardware.


\subsection{Enabling exploration with domain knowledge}
\label{sec:exploration}
A key issue in using reinforcement learning for learning precision in-hand manipulation skills is that a random exploratory action can easily disturb the stability of the object held in a precision grasp, causing it to be dropped. This difficulty is particularly acute for finger-gaiting, which requires fingertips to break contact with the object and transition between different grasps, involving different fingertips, all while re-orienting the object. Intuitively, the likelihood of selecting a sequence of random actions that can accomplish this feat and obtain a useful reward signal is very low.


For a policy to learn finger-gaiting, it must encounter these diverse grasps within its training samples so that the policy's action distributions can improve at these states.
Consider taking a sequence of random actions starting from a stable $l$-finger grasp. While it is possible to reach a stable grasp with an additional finger in contact (if available), it is more likely to lose one finger contact, then another and so on until the object is dropped. Over multiple trials, we can expect to encounter most combinations of $l-1$ grasps. In this setting, it can be argued that starting from a stable grasp with all $m$ fingers in contact leads to maximum exploration. Interestingly, as we will demonstrate in Sec \ref{sec:expts}, we found this to be insufficient.

Our insight is to observe that through domain knowledge we are already aware of the states that a sufficiently exploratory policy might visit. Using domain knowledge in designing initial distributions is a known way of improving sample complexity \cite{Kakade2003-ja}\cite{De_Farias2003-ai}. Thus, we use our knowledge of relevant states in designing the initial states used for episode rollouts and show that it is critical for learning precision finger-gaiting and finger-pivoting. 

We propose sampling sufficiently-varied stable grasps relevant to re-orienting the object about the desired axis and use them as initial states for collecting training trajectories. These grasps must be well distributed in terms of number of contacts, contact positions relative to the object, and object poses relevant to the task. To this end, we first initialize the object in an random pose and then sample fingertip positions until we find a stable grasp as described in Stable Grasp Sampling (SGS) in Alg. \ref{algo:sampling}.

\begin{figure}[t]
    \centering
    \includegraphics[width=\columnwidth]{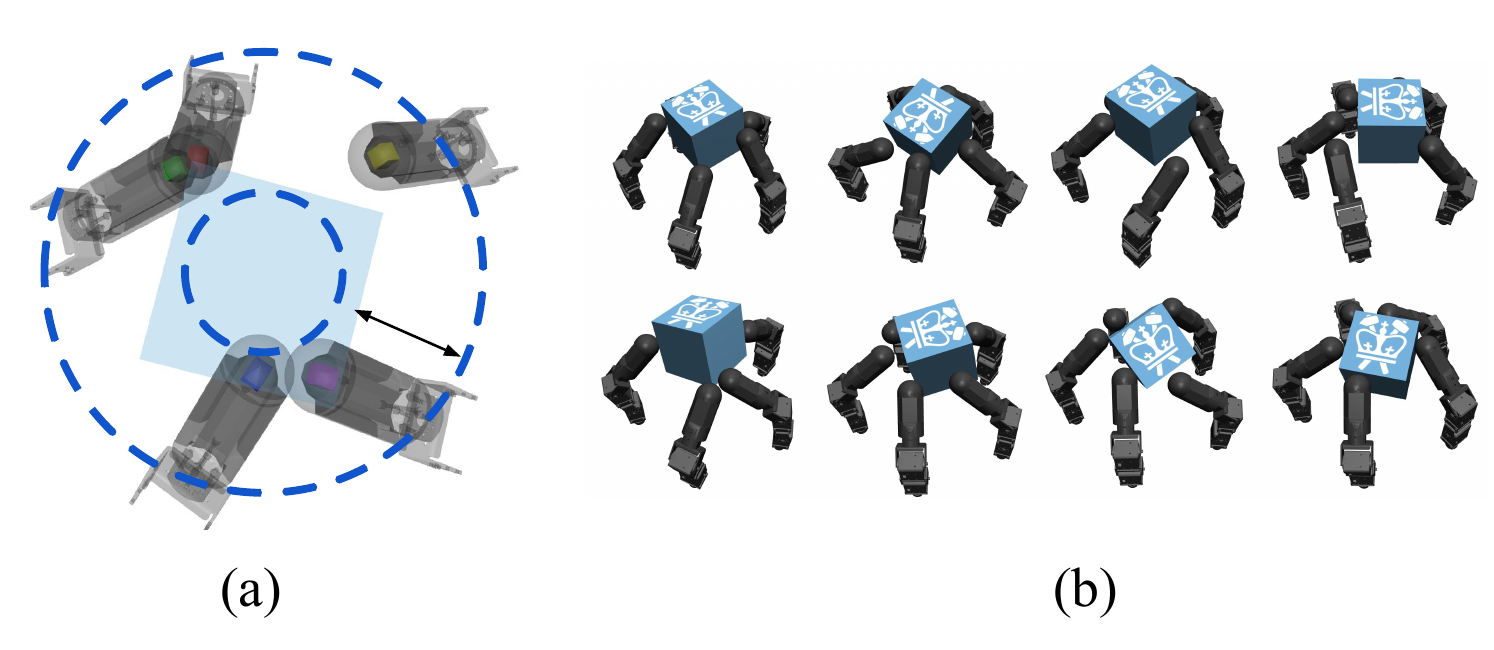}
    \caption{(a) Sampling fingertips around the object. (b) Diverse relevant initial grasps sampled for efficient exploration.}
    \label{fig:init_dist}
    \vspace{-5pt}
\end{figure}

\begin{algorithm}[b]
    \caption{Stable Grasp Sampling (SGS)} \label{algo:sampling}
    \textbf{Input:}$\rho_{obj}$, $\rho_{hand}$, $t_s$, $n_{c,\min}$ \Comment{object pose distribution, hand pose distribution, simulation settling time, minimum number of contacts} \\
    \textbf{Output:} $\mathbf{s}_{g}$ \Comment{simulator state of the sampled grasp} 
    \begin{algorithmic}[1]
    \Repeat
    \State Sample object and hand pose: $\boldsymbol{x}_s$ $\sim$ $\rho_{obj}$, $\boldsymbol{q}_s$ $\sim$ $\rho_{hand}$
    \State Set object pose in the simulator with $\boldsymbol{x}_s$
    \State Set joint positions and controller set-points with $\boldsymbol{q}_s$
    \State Step the simulation forward by $t_s$ seconds
    \State Find number of fingertips in contact with object, $n_c$
    \Until{$n_c \geq n_{c,\min}$}
    \State Save simulator state as $\boldsymbol{s}_{g}$
    \end{algorithmic}
\end{algorithm}

In SGS, we first sample an object pose and a hand pose, then update the simulator with the sampled poses towards obtaining a grasp. We advance the simulation for a short duration, $t_s$, to let any transients die down. If the object has settled into a grasp with at least two contacts, the pose is used for collecting training trajectories. Note that the fingertips could be overlapping with the object or with each other as we do not explicitly check this. However, due to the soft-contact model used by the simulator (MuJoCo \cite{Todorov2012-xe}) the inter-penetrations are resolved during simulation.  An illustrative set of grasps sampled by SGS are shown in Fig \ref{fig:init_dist}b.

To sample the hand pose, we start by sampling fingertip locations within an annulus centered on and partially overlaps with the object (Fig \ref{fig:init_dist}a). Thus, the probabilities of each fingertip making contact with the object and of staying free are roughly the same. With this procedure, not only do we find stable grasps relevant to finger-gaiting and finger-pivoting, we improve the likelihood of discovering them, thus minimizing training wall-clock time.

\section{Experiments and Results}
For evaluating our method, we focus on learning precision in-hand re-orientation about the z- and x- axes for a range of regular object shapes.
(The y-axis is similar to x-, given the symmetry of our hand model.) Our object set, which consists of a cylinder, sphere, icosahedron, dodecahedron and cube, is designed so that we have objects of varying difficulty with the sphere and cube being the easiest and hardest, respectively. For training, we use PPO \cite{Schulman2017-fr}. We chose PPO over other state-of-the-art methods such as SAC primarily for training stability .

For the following analysis, we use z-axis re-orientation as a case study. In addition to the above, we also train z-axis re-orientation policies without assuming joint set-point feedback $\mathbf{q}_d$. For all these policies, we study their robustness properties by adding noise and also by applying perturbation forces on the object (Sec \ref{sec:robust}). We also study the zero-shot generalization properties of these policies (Sec \ref{sec:gen}). Finally, through ablation studies we present a detailed analysis ascertaining the importance of different components of feedback for achieving finger-pivoting (Sec \ref{sec:fb_study}).

We note that, in simulation, the combination of $\mathbf{q}_d$ and $\mathbf{q}$ can be considered a good proxy for torque, since simulated controllers have stable and known stiffness. However, this feature might not transfer to a real hand, where transmissions exhibit friction, stiction and other hard to model effects. We thus evaluate our policies both with and without joint set-point observations.
\begin{figure*}[ht!]
    \centering
    \includegraphics[clip, width=0.75\textwidth]{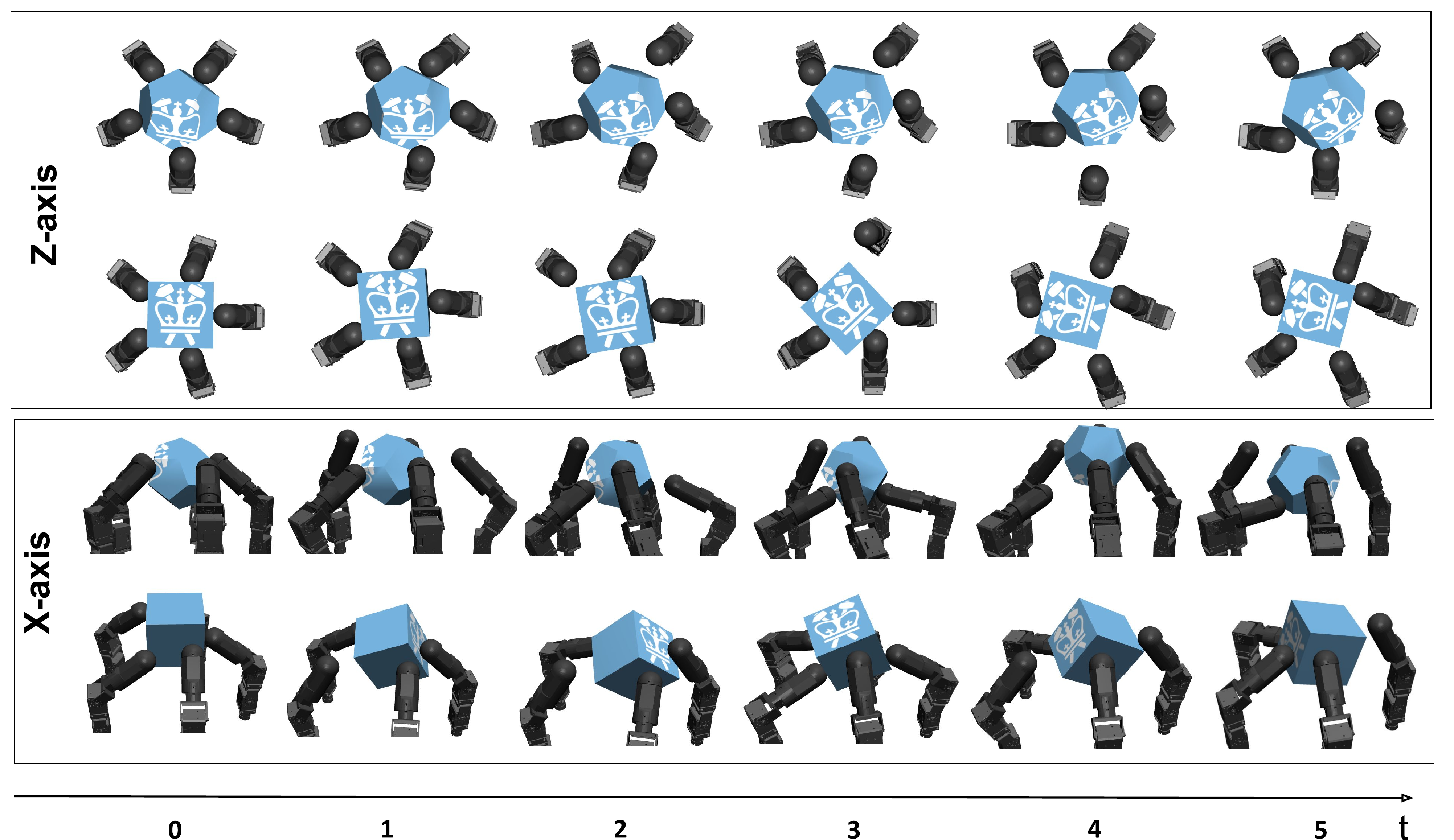}
    \caption{Finger-gaiting and finger-pivoting our policies achieve to re-orient about z-axis and x-axis respectively. Key frames are shown for two objects, dodecahedron and cube. }
    \label{fig:gaitframes}
\end{figure*}
\subsection{Learning finger-gaiting manipulation}
\label{sec:expts}
\begin{figure}[t]
    \centering
    \includegraphics[clip, width=0.9\columnwidth]{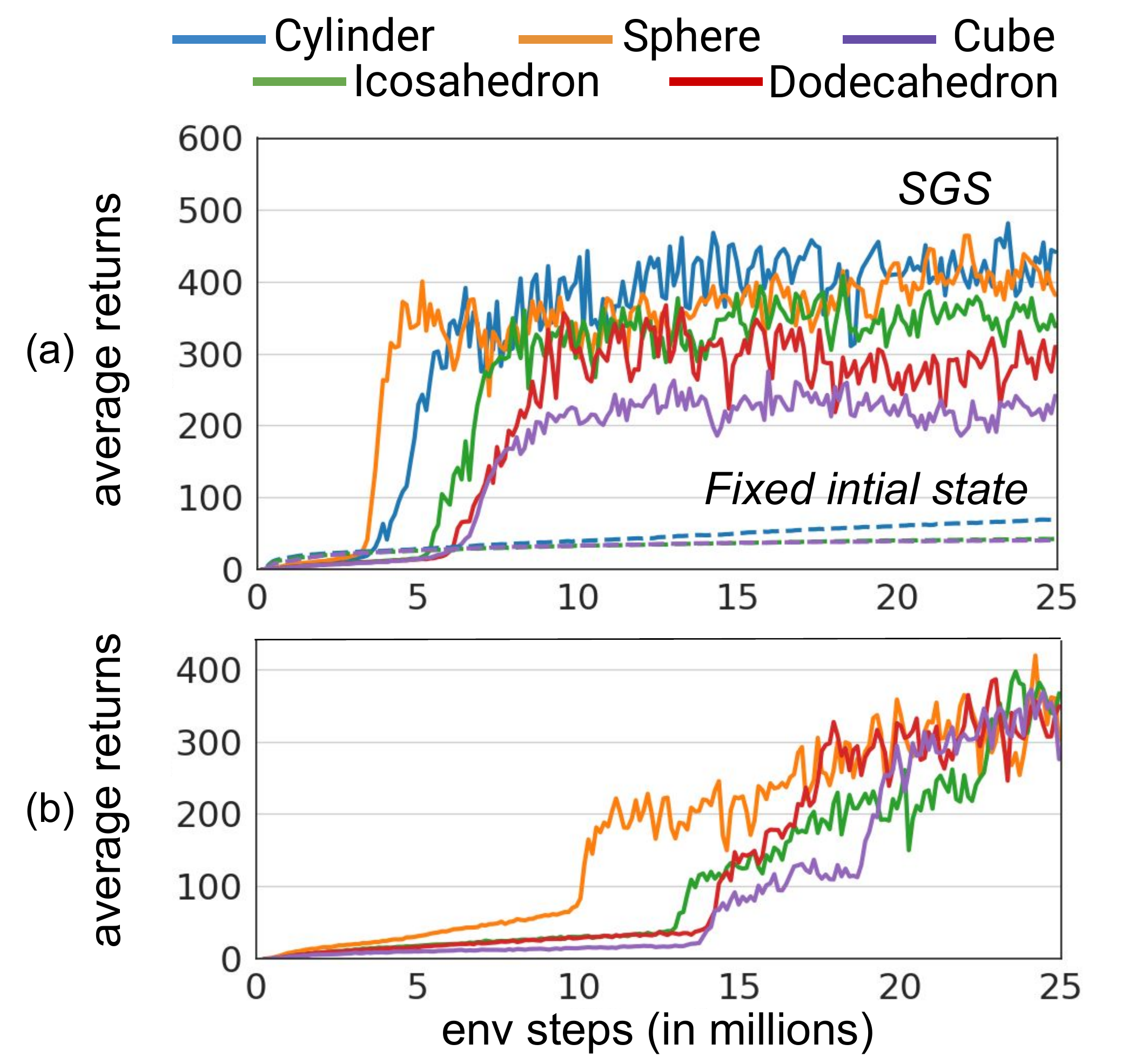}
    \caption{Average returns for (a) z-axis re-orientation and (b) x-axis re-orientation. Learning with wide range of initial grasps sampled via SGS succeeds, while using a fixed initial state fails.}
    \label{fig:train_return}
    \vspace{-10pt}
\end{figure}

Fig \ref{fig:train_return}a shows the learning curves for object re-orientation about the z-axis for a range of objects from using our method of sampling stable initial grasps to improve exploration. We also show learning curves using a fixed initial state (grasp with all fingers) for representative objects. First, we notice that the latter approach does not succeed. These policies only achieve small re-orientation via in-grasp manipulation and drop the object after maximum re-orientation achievable without breaking contacts. 

However, when using a wide initial distribution of grasps (sampled via SGS), the policies learn finger-gaiting and achieve continuous re-orientation of the object with significantly higher returns. With our approach, we also learn finger-pivoting for re-orientation about the x-axis, with learning curves shown in Fig \ref{fig:train_return}b. Thus, we empirically see that using a wide initial distribution consisting of relevant grasps is critical for learning continuous in-hand re-orientation and that our method results in superior sample-complexity over the state-of-the-art i.e PPO without the use of initial state distribution. Fig \ref{fig:gaitframes} shows our finger-gaiting and finger-pivoting policies performing continuous object re-orientation about z-axis and x-axis respectively.

As expected, difficulty of rotating the objects increases as we consider objects of lower rotational symmetry from sphere to cube. In the training curves in Fig \ref{fig:train_return}, we can observe this trend not only in the final returns achieved by the respective policies, but also in the number of samples required to learn continuous re-orientation.

We also successfully learn policies for in-hand re-orientation without joint set-point position feedback, but these policies achieve slightly lower returns. However, they may have interesting consequences for generalization as we will discuss in Sec \ref{sec:gen}.

\subsection{Robustness}
\label{sec:robust}
\begin{figure}[t]
    \centering
    \includegraphics[clip,width=0.8\columnwidth]{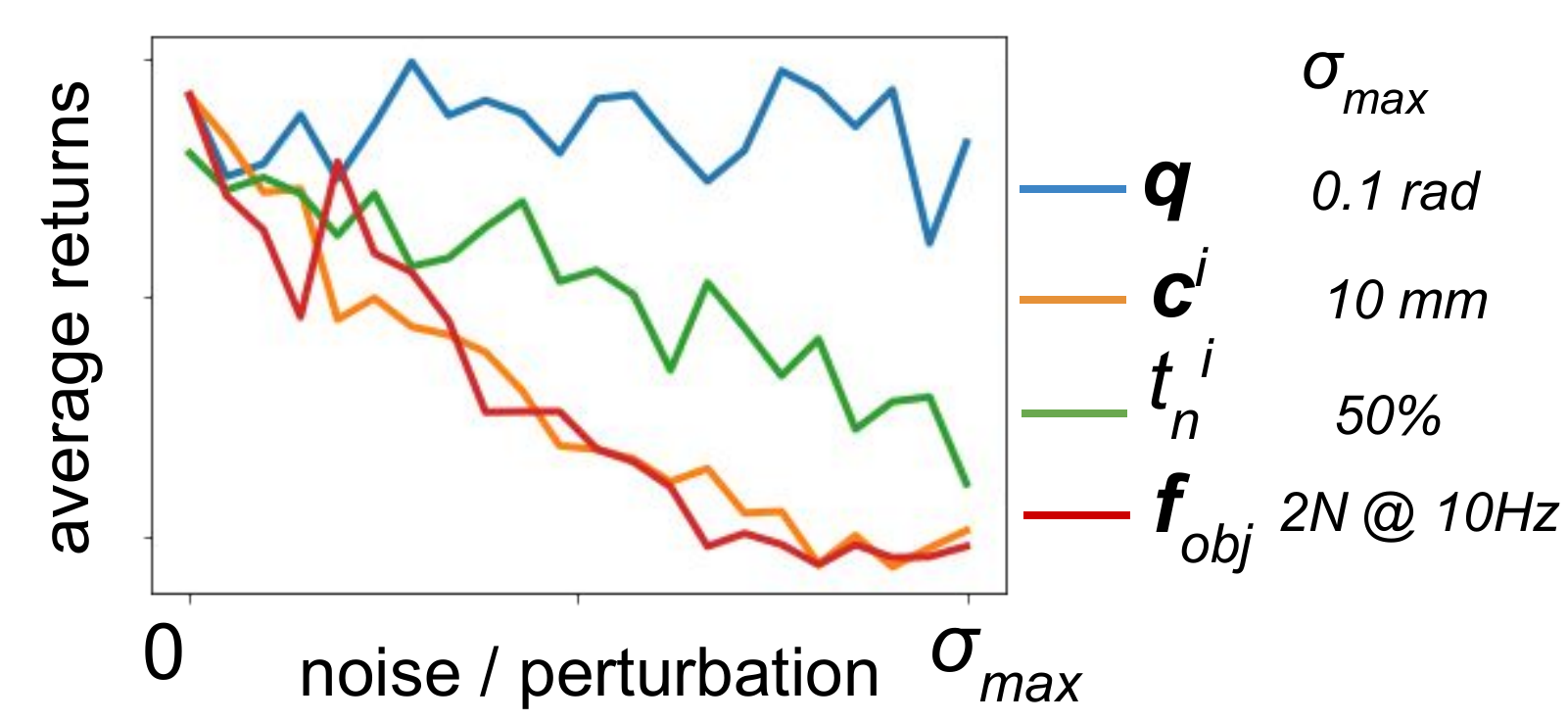}
    \caption{Robustness of our policies with increasing sensor noise and perturbation forces on the object.}
    \label{fig:robustness}
\end{figure}

Fig. \ref{fig:robustness} shows the performance of our policy for the most difficult object in our set (cube) as we artificially add white noise with increasing variance to different sensors' feedback. We also increasingly add perturbation forces on the object. Overall, we notice that our policies are robust to noise and perturbation forces of magnitudes that can be expected on a real hand.

In particular, our policies show little drop in performance for noise in joint positions, but are more sensitive to noise in contact feedback. Nevertheless, they are still robust, and achieve high returns even at $5$mm error in contact position and $25\%$ error in contact force.  Interestingly, for noise in contact position, we found that drop in performance arises indirectly through the error in contact normal $\hat{\mathbf{t}}_n^i$ (computed from contact position $\mathbf{c}_n^i$). As for perturbation forces on the object, we observe high returns even for high perturbation forces ($1$N) equivalent to the weight of our objects. Our policies are robust event without joint-setpoint $\mathbf{q}_d$ feedback with similar robustness profiles.

\subsection{Generalization}
\label{sec:gen}
We study generalization properties of our policies by evaluating it on  different objects in the object set. We consider the transfer score, which is the ratio $R_{ij}/R_{ii}$ where $R_{ij}$ is the average returns obtained when evaluating the policy learned with object $i$ on object $j$. 

\begin{figure}
    \centering
    \includegraphics[clip, width=\columnwidth]{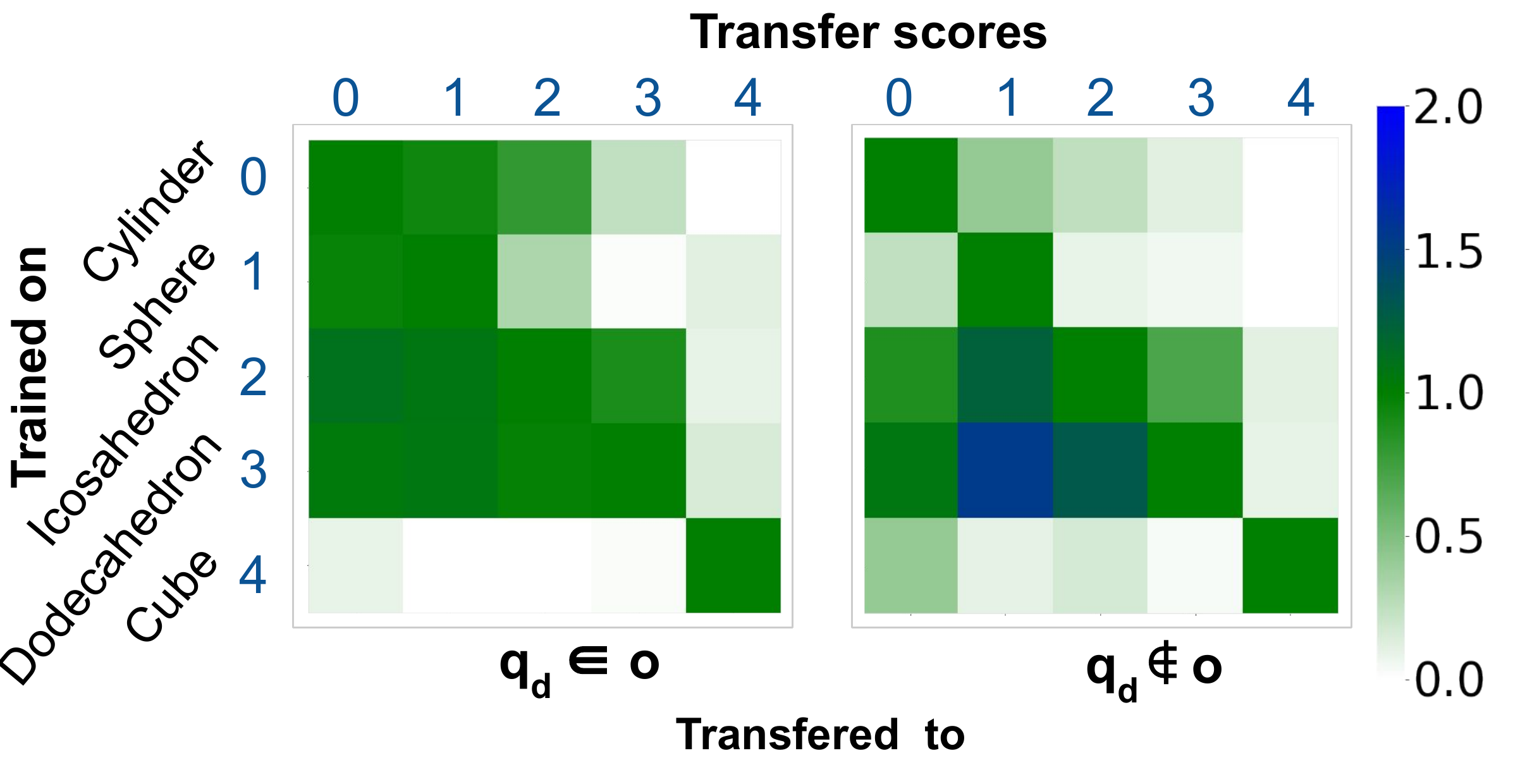}
    \caption{Cross transfer scores for policies with and without $\mathbf{q}_d$ in feedback.}
    \label{fig:cross_transfer_a}
    \vspace{-5pt}
\end{figure}

Fig. \ref{fig:cross_transfer_a} shows the cross transfer performance for policies trained with all feedback. We note that the policy trained on the sphere transfers to the cylinder and vice versa. Also, the policies trained on icosahedron and dodecahedron transfer well between themselves and also perform well on sphere and cylinder. Interestingly, the policy trained on the cube does not transfer well to the other objects. When not using joint set-point position feedback $\mathbf{q}_d$, the policy learned on the cube transfers to more objects. With no way to infer motor forces, the policy potentially learns to rely more on contact feedback which aids generalization.

\subsection{Observations on feedback}
\label{sec:fb_study}
\begin{figure}
    \centering
    \includegraphics[width=0.9\columnwidth]{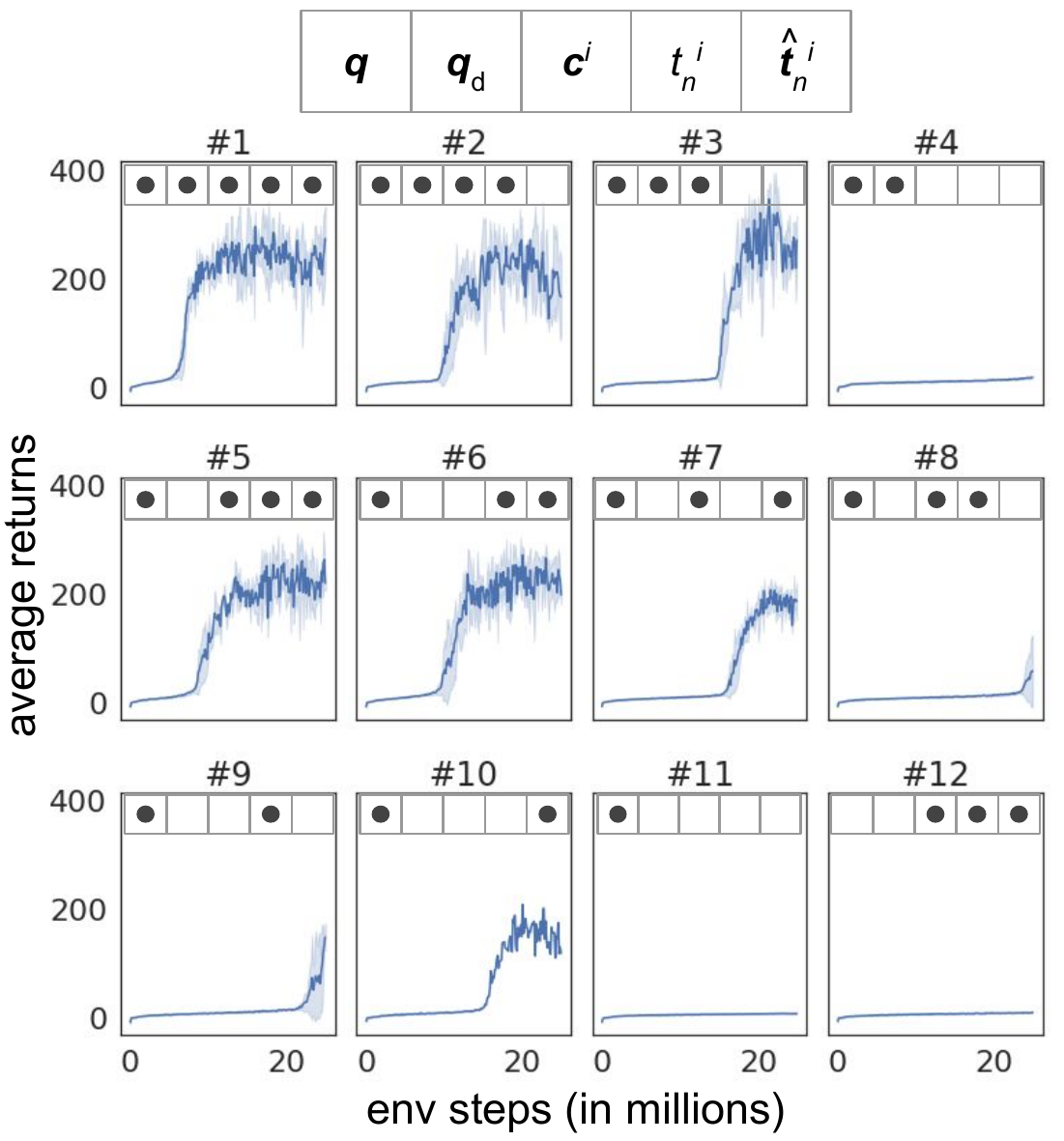}
    \caption{Ablations holding out different components of feedback. For each experiment, dots in the observation vector shown above the training curve indicate which of the components of the observation vector are provided to the policy.}
    \label{fig:fb_study}
    \vspace{-5pt}
\end{figure}

While our work provides some insight w.r.t the important components of our feedback through our robustness and generalization results, many interesting questions remain. We are particularly interested to discover what aspects matter most in contact feedback. To answer such questions, we run a series of ablations holding out different components.  For this, we again consider learning finger-gaiting on the cube as shown in Fig \ref{fig:fb_study}. 

Based on this ablation study, we can make a number of observations. As expected, contact feedback is essential for learning in-hand re-orientation via finger-gaiting; the policy does not learn finger-gaiting with just proprioceptive feedback (\#4). More interesting, and also more surprising, is that explicitly computing contact normal $\mathbf{t}_n^i$ and providing it as feedback is critical when excluding  joint position set-point $\mathbf{q}_d$ (\#6 to \#10). In fact, the policy learns finger-gaiting with just contact normal and joint position feedback (\#10). However, while not critical, contact position and force feedback are still beneficial as they improve sample efficiency (\#6, \#7).

\section{Conclusion}
In this paper, we focus on the problem of learning in-hand manipulation policies that can achieve large-angle object re-orientation via finger-gaiting. To facilitate future deployment in real scenarios, we restrict ourselves to using sensing modalities intrinsic to the hand, such as touch and proprioception, with no external vision or tracking sensor providing object-specific information. Furthermore, we aim for policies that can achieve manipulation skills without using a palm or other surfaces for passive support, and which instead need to maintain the object in a stable grasp.

A critical component of our approach is the use of appropriate initial state distributions during training, used to alleviate the intrinsic instability of precision grasping. We also decompose the manipulation problem into axis-specific rotation policies in the hand coordinate frame, allowing for object-agnostic policies. Combining these, we are able to achieve the desired skills in a simulated environment, the first instance in the literature of such policies being successfully trained with intrinsic sensor data.

We consider this work to be a useful step towards future sim-to-real transfer. To this end, we engage in an exhaustive empirical analysis of the role that each sensing modality plays in enabling our manipulation skills. Specifically, we show that tactile feedback in addition to proprioceptive sensing is critical in enabling such performance. Finally, our analysis of the policies shows that they generalize to novel objects and are also sufficiently robust to force perturbations and sensing noise, suggesting the possibility of future sim-to-real transfer. 

\addtolength{\textheight}{0cm}
\printbibliography

@INPROCEEDINGS{Okamura2000-in,
  title     = "An overview of dexterous manipulation",
  booktitle = "Proceedings 2000 {ICRA}. Millennium Conference. {IEEE}
               International Conference on Robotics and Automation. Symposia
               Proceedings (Cat. {No.00CH37065})",
  author    = "Okamura, A M and Smaby, N and Cutkosky, M R",
  volume    =  1,
  pages     = "255--262 vol.1",
  month     =  apr,
  year      =  2000
}

@INPROCEEDINGS{Omata1996-hp,
  title     = "Regrasps by a multifingered hand based on primitives",
  booktitle = "Proceedings of {IEEE} International Conference on Robotics and
               Automation",
  author    = "Omata, T and Farooqi, M A",
  volume    =  3,
  pages     = "2774--2780 vol.3",
  month     =  apr,
  year      =  1996
}

@INPROCEEDINGS{Leveroni1996-iy,
  title     = "Reorienting Objects with a Robot Hand Using Grasp Gaits",
  booktitle = "Robotics Research",
  author    = "Leveroni, Susanna and Salisbury, Kenneth",
  publisher = "Springer London",
  pages     = "39--51",
  year      =  1996
}

@INPROCEEDINGS{Han1998-xj,
  title     = "Dextrous manipulation by rolling and finger gaiting",
  booktitle = "Proceedings. 1998 {IEEE} International Conference on Robotics
               and Automation (Cat. {No.98CH36146})",
  author    = "Han, L and Trinkle, J C",
  volume    =  1,
  pages     = "730--735 vol.1",
  month     =  may,
  year      =  1998
}

@ARTICLE{Michelman1998-ye,
  title    = "Precision object manipulation with a multifingered robot hand",
  author   = "Michelman, P",
  journal  = "IEEE Trans. Rob. Autom.",
  volume   =  14,
  number   =  1,
  pages    = "105--113",
  month    =  feb,
  year     =  1998
}

@INPROCEEDINGS{Platt2004-nr,
  title     = "Manipulation gaits: sequences of grasp control tasks",
  booktitle = "{IEEE} International Conference on Robotics and Automation,
               2004. Proceedings. {ICRA} '04. 2004",
  author    = "Platt, R and Fagg, A H and Grupen, R A",
  volume    =  1,
  pages     = "801--806 Vol.1",
  month     =  apr,
  year      =  2004
}

@INPROCEEDINGS{Tassa2012-hx,
  title     = "Synthesis and stabilization of complex behaviors through online
               trajectory optimization",
  booktitle = "2012 {IEEE/RSJ} International Conference on Intelligent Robots
               and Systems",
  author    = "Tassa, Y and Erez, T and Todorov, E",
  pages     = "4906--4913",
  month     =  oct,
  year      =  2012
}

@article{Fan2017-vz,
title = {Real-Time Finger Gaits Planning for Dexterous Manipulation**This project was supported by FANUC Corporation},
journal = {IFAC-PapersOnLine},
volume = {50},
number = {1},
pages = {12765-12772},
year = {2017},
note = {20th IFAC World Congress},
issn = {2405-8963},
%doi = {https://doi.org/10.1016/j.ifacol.2017.08.1831},
%url = {https://www.sciencedirect.com/science/article/pii/S2405896317324540},
author = {Yongxiang Fan and Wei Gao and Wenjie Chen and Masayoshi Tomizuka},
keywords = {dexterous manipulation, finger gaits planning, multi-fingered hands, autonomous robotic systems, robotics technology, modeling for control optimization, real-time control},
abstract = {Dexterous manipulation has broad potential applications in assembly lines, warehouses and agriculture, and so on. To perform large-scale, complicated manipulation tasks, a multi-fingered robotic hand sometimes has to sequentially adjust its grasping status, i.e. the finger gaits, to deal with the workspace limits and object stability. However, realizing finger gaits planning in dexterous manipulation is challenging due to the involved hybrid dynamics, complicated grasp quality metrics, and uncertainties during the finger gaiting. In this paper, a dual-stage optimization based controller is proposed to handle these challenges. First, a velocity-level finger gaits planner is introduced by combining object grasp quality with hand kinematic limitations. The proposed finger gaits planner is computationally efficient and can be solved in real-time. Second, a manipulation controller using force optimization is presented. To deal with mass uncertainties and external disturbances, a modified impedance control is integrated into the manipulation controller. The dual-stage controller does not require the shape of the object, nor does it rely on expensive 3D/6D tactile sensors. Simulation results verify the efficacy of the proposed dual-stage controller.}
}

@ARTICLE{Li2019-kj,
%   title         = "Learning Hierarchical Control for Robust {In-Hand}
%                   Manipulation",
%   author        = "Li, Tingguang and Srinivasan, Krishnan and Meng, Max Qing-Hu
%                   and Yuan, Wenzhen and Bohg, Jeannette",
%   abstract      = "Robotic in-hand manipulation has been a long-standing
%                   challenge due to the complexity of modelling hand and object
%                   in contact and of coordinating finger motion for complex
%                   manipulation sequences. To address these challenges, the
%                   majority of prior work has either focused on model-based,
%                   low-level controllers or on model-free deep reinforcement
%                   learning that each have their own limitations. We propose a
%                   hierarchical method that relies on traditional, model-based
%                   controllers on the low-level and learned policies on the
%                   mid-level. The low-level controllers can robustly execute
%                   different manipulation primitives (reposing, sliding,
%                   flipping). The mid-level policy orchestrates these
%                   primitives. We extensively evaluate our approach in
%                   simulation with a 3-fingered hand that controls three
%                   degrees of freedom of elongated objects. We show that our
%                   approach can move objects between almost all the possible
%                   poses in the workspace while keeping them firmly grasped. We
%                   also show that our approach is robust to inaccuracies in the
%                   object models and to observation noise. Finally, we show how
%                   our approach generalizes to objects of other shapes.",
%   month         =  oct,
%   year          =  2019,
%   keywords      = "In-hand Manipulation;In-hand manipulation CoRL 2021",
%   archivePrefix = "arXiv",
%   primaryClass  = "cs.RO",
%   eprint        = "1910.10985"
% }

@ARTICLE{Sundaralingam2018-zw,
%   title         = "Geometric {In-Hand} Regrasp Planning: Alternating
%                   Optimization of Finger Gaits and {In-Grasp} Manipulation",
%   author        = "Sundaralingam, Balakumar and Hermans, Tucker",
%   abstract      = "This paper explores the problem of autonomous, in-hand
%                   regrasping--the problem of moving from an initial grasp on
%                   an object to a desired grasp using the dexterity of a
%                   robot's fingers. We propose a planner for this problem which
%                   alternates between finger gaiting, and in-grasp
%                   manipulation. Finger gaiting enables the robot to move a
%                   single finger to a new contact location on the object, while
%                   the remaining fingers stably hold the object. In-grasp
%                   manipulation moves the object to a new pose relative to the
%                   robot's palm, while maintaining the contact locations
%                   between the hand and object. Given the object's geometry (as
%                   a mesh), the hand's kinematic structure, and the initial and
%                   desired grasps, we plan a sequence of finger gaits and
%                   object reposing actions to reach the desired grasp without
%                   dropping the object. We propose an optimization based
%                   approach and report in-hand regrasping plans for 5 objects
%                   over 5 in-hand regrasp goals each. The plans generated by
%                   our planner are collision free and guarantee kinematic
%                   feasibility.",
%   month         =  apr,
%   year          =  2018,
%   keywords      = "In-hand Manipulation;In-hand manipulation CoRL 2021",
%   archivePrefix = "arXiv",
%   primaryClass  = "cs.RO",
%   eprint        = "1804.04292",
%   journal       = ""
% }

@INPROCEEDINGS{Van_Hoof2015-wm,
  title     = "Learning robot in-hand manipulation with tactile features",
  booktitle = "2015 {IEEE-RAS} 15th International Conference on Humanoid Robots
               (Humanoids)",
  author    = "van Hoof, Herke and Hermans, Tucker and Neumann, Gerhard and
               Peters, Jan",
  pages     = "121--127",
  month     =  nov,
  year      =  2015
}

@article{OpenAI2018-bx,
author = {OpenAI: Marcin Andrychowicz and Bowen Baker and Maciek Chociej and Rafal Józefowicz and Bob McGrew and Jakub Pachocki and Arthur Petron and Matthias Plappert and Glenn Powell and Alex Ray and Jonas Schneider and Szymon Sidor and Josh Tobin and Peter Welinder and Lilian Weng and Wojciech Zaremba},
title ={Learning dexterous in-hand manipulation},
journal = {The International Journal of Robotics Research},
volume = {39},
number = {1},
pages = {3-20},
year = {2020},
doi = {10.1177/0278364919887447},

% URL = { 
%         https://doi.org/10.1177/0278364919887447
    
% },
% eprint = { 
%         https://doi.org/10.1177/0278364919887447
    
% }
% ,
    abstract = { We use reinforcement learning (RL) to learn dexterous in-hand manipulation policies that can perform vision-based object reorientation on a physical Shadow Dexterous Hand. The training is performed in a simulated environment in which we randomize many of the physical properties of the system such as friction coefficients and an object’s appearance. Our policies transfer to the physical robot despite being trained entirely in simulation. Our method does not rely on any human demonstrations, but many behaviors found in human manipulation emerge naturally, including finger gaiting, multi-finger coordination, and the controlled use of gravity. Our results were obtained using the same distributed RL system that was used to train OpenAI Five. We also include a video of our results: https://youtu.be/jwSbzNHGflM. }
}

@ARTICLE{Zhu2019-rk,
%   author    = {Henry Zhu and
%               Abhishek Gupta and
%               Aravind Rajeswaran and
%               Sergey Levine and
%               Vikash Kumar},
%   title     = {Dexterous Manipulation with Deep Reinforcement Learning: Efficient,
%               General, and Low-Cost},
%   journal   = {CoRR},
%   volume    = {abs/1810.06045},
%   year      = {2018},
%   url       = {http://arxiv.org/abs/1810.06045},
%   archivePrefix = {arXiv},
%   eprint    = {1810.06045},
%   timestamp = {Thu, 20 Dec 2018 16:30:14 +0100},
%   biburl    = {https://dblp.org/rec/journals/corr/abs-1810-06045.bib},
%   bibsource = {dblp computer science bibliography, https://dblp.org}
% }

@ARTICLE{Rajeswaran2017-fw,
%   title         = "Learning Complex Dexterous Manipulation with Deep
%                   Reinforcement Learning and Demonstrations",
%   author        = "Rajeswaran, Aravind and Kumar, Vikash and Gupta, Abhishek
%                   and Vezzani, Giulia and Schulman, John and Todorov, Emanuel
%                   and Levine, Sergey",
%   abstract      = "Dexterous multi-fingered hands are extremely versatile and
%                   provide a generic way to perform a multitude of tasks in
%                   human-centric environments. However, effectively controlling
%                   them remains challenging due to their high dimensionality
%                   and large number of potential contacts. Deep reinforcement
%                   learning (DRL) provides a model-agnostic approach to control
%                   complex dynamical systems, but has not been shown to scale
%                   to high-dimensional dexterous manipulation. Furthermore,
%                   deployment of DRL on physical systems remains challenging
%                   due to sample inefficiency. Consequently, the success of DRL
%                   in robotics has thus far been limited to simpler
%                   manipulators and tasks. In this work, we show that
%                   model-free DRL can effectively scale up to complex
%                   manipulation tasks with a high-dimensional 24-DoF hand, and
%                   solve them from scratch in simulated experiments.
%                   Furthermore, with the use of a small number of human
%                   demonstrations, the sample complexity can be significantly
%                   reduced, which enables learning with sample sizes equivalent
%                   to a few hours of robot experience. The use of
%                   demonstrations result in policies that exhibit very natural
%                   movements and, surprisingly, are also substantially more
%                   robust.",
%   month         =  sep,
%   year          =  2017,
%   keywords      = "In-hand Manipulation;Dexterous Manipulation;In-hand
%                   manipulation CoRL 2021",
%   archivePrefix = "arXiv",
%   primaryClass  = "cs.LG",
%   eprint        = "1709.10087"
% }

@INPROCEEDINGS{Shi2020-dd,
  author={Shi, Fan and Homberger, Timon and Lee, Joonho and Miki, Takahiro and Zhao, Moju and Farshidian, Farbod and Okada, Kei and Inaba, Masayuki and Hutter, Marco},
  booktitle={2021 IEEE International Conference on Robotics and Automation (ICRA)}, 
  title={Circus ANYmal: A Quadruped Learning Dexterous Manipulation with Its Limbs}, 
  year={2021},
  volume={},
  number={},
  pages={2316-2323},
  doi={10.1109/ICRA48506.2021.9561926}}

@INPROCEEDINGS{Morgan2021-ny,
  author={Morgan, Andrew S. and Nandha, Daljeet and Chalvatzaki, Georgia and D’Eramo, Carlo and Dollar, Aaron M. and Peters, Jan},
  booktitle={2021 IEEE International Conference on Robotics and Automation (ICRA)}, 
  title={Model Predictive Actor-Critic: Accelerating Robot Skill Acquisition with Deep Reinforcement Learning}, 
  year={2021},
  volume={},
  number={},
  pages={6672-6678},
  doi={10.1109/ICRA48506.2021.9561298}}

@article{Radosavovic2020-av,
  title={State-Only Imitation Learning for Dexterous Manipulation},
  author={Ilija Radosavovic and Xiaolong Wang and Lerrel Pinto and Jitendra Malik},
  journal={2021 IEEE/RSJ International Conference on Intelligent Robots and Systems (IROS)},
  year={2021},
  pages={7865-7871}
}

@INPROCEEDINGS{Schulman2017-fr,
  title         = "Proximal Policy Optimization Algorithms",
  author        = "Schulman, John and Wolski, Filip and Dhariwal, Prafulla and Radford, Alec and Klimov, Oleg",
  month         =  jul,
  year          =  2017,
  archivePrefix = "arXiv",
  primaryClass  = "cs.LG",
  eprint        = "1707.06347"
}

@INPROCEEDINGS{Todorov2012-xe,
  title     = "{MuJoCo}: A physics engine for model-based control",
  booktitle = "2012 {IEEE/RSJ} International Conference on Intelligent Robots
               and Systems",
  author    = "Todorov, Emanuel and Erez, Tom and Tassa, Yuval",
  pages     = "5026--5033",
  month     =  oct,
  year      =  2012
}

@PHDTHESIS{Kakade2003-ja,
  title     = "On the sample complexity of reinforcement learning",
  author    = "Kakade, Sham Machandranath",
  publisher = "University College London (United Kingdom). ProQuest
               Dissertations Publishing",
  year      =  2003,
  address   = "Ann Arbor, United States",
  school    = "University of London, University College London (United Kingdom)",
  language  = "en"
}

@ARTICLE{De_Farias2003-ai,
  title     = "The linear programming approach to approximate dynamic
               programming",
  author    = "de Farias, D P and Van Roy, B",
  journal   = "Oper. Res.",
  publisher = "Institute for Operations Research and the Management Sciences
               (INFORMS)",
  volume    =  51,
  number    =  6,
  pages     = "850--865",
  month     =  dec,
  year      =  2003
}

@ARTICLE{Nagabandi2020-hz,
  title     = "Deep dynamics models for learning dexterous manipulation",
  author    = "Nagabandi, A and Konolige, K and Levine, S and {others}",
  journal   = "Conference on Robot",
  publisher = "proceedings.mlr.press",
  year      =  2020
}

@ARTICLE{Li2020-mc,
  title    = "A Review of Tactile Information: Perception and Action Through
              Touch",
  author   = "Li, Qiang and Kroemer, Oliver and Su, Zhe and Veiga, Filipe
              Fernandes and Kaboli, Mohsen and Ritter, Helge Joachim",
  journal  = "IEEE Trans. Rob.",
  volume   =  36,
  number   =  6,
  pages    = "1619--1634",
  month    =  dec,
  year     =  2020
}

@ARTICLE{Veiga2020-zm,
  title    = "Hierarchical {Tactile-Based} Control Decomposition of Dexterous
              {In-Hand} Manipulation Tasks",
  author   = "Veiga, Filipe and Akrour, Riad and Peters, Jan and ,",
  journal  = "Front Robot AI",
  volume   =  7,
  pages    = "521448",
  month    =  nov,
  year     =  2020,
  language = "en"
}

@ARTICLE{Melnik2021openAItactile,
  title    = "Using Tactile Sensing to Improve the Sample Efficiency and
              Performance of Deep Deterministic Policy Gradients for Simulated
              {In-Hand} Manipulation Tasks",
  author   = "Melnik, Andrew and Lach, Luca and Plappert, Matthias and
              Korthals, Timo and Haschke, Robert and Ritter, Helge",
  journal  = "Front Robot AI",
  volume   =  8,
  pages    = "538773",
  month    =  jun,
  year     =  2021,
  language = "en"
}

@INPROCEEDINGS{Saut2007-su,
  title     = "Dexterous manipulation planning using probabilistic roadmaps in
               continuous grasp subspaces",
  booktitle = "2007 {IEEE/RSJ} International Conference on Intelligent Robots
               and Systems",
  author    = "Saut, Jean-Philippe and Sahbani, Anis and El-Khoury, Sahar and
               Perdereau, Veronique",
  pages     = "2907--2912",
  month     =  oct,
  year      =  2007
}

@inproceedings{
chen2021a,
title={A Simple Method for Complex In-hand Manipulation},
author={Tao Chen and Jie Xu and Pulkit Agrawal},
booktitle={5th Annual Conference on Robot Learning },
year={2021},
%url={https://openreview.net/forum?id=7uSBJDoP7tY}
}

\end{document}